\documentclass[letterpaper, 10 pt, conference]{ieeeconf}  
\IEEEoverridecommandlockouts                              
\usepackage{amsmath,amssymb,amsfonts}
\usepackage{algorithmic}
\usepackage[ruled,vlined,linesnumbered]{algorithm2e}    
\usepackage{array}
\usepackage[caption=false,font=footnotesize]{subfig}
\usepackage{textcomp}
\usepackage{stfloats}
\usepackage{url}
\usepackage{verbatim}
\usepackage{graphicx}
\usepackage{comment}
\usepackage{cite}
\usepackage{enumerate}
\usepackage{xcolor}
\usepackage{xfrac}
\usepackage{tabularx}
\usepackage{hyperref}
\usepackage{pdflscape}
\hypersetup{
    colorlinks=true,
    linkcolor=blue,
    filecolor=magenta,      
    urlcolor=cyan,
    pdftitle={Overleaf Example},
    pdfpagemode=FullScreen,
    }

\title{\LARGE \bf
Resilient Fleet Management for Energy-Aware Intra-Factory Logistics
}

\author{Mithun Goutham and Stephanie Stockar
\thanks{Mithun Goutham and Stephanie Stockar are with the Department of Mechanical and Aerospace Engineering,
        The Ohio State University, Columbus, OH 43210, USA
        {\tt\small goutham.1@osu.edu}}
\thanks{
        This work was presented at the 2024 American Control Conference and is published under DOI: 10.23919/ACC60939.2024.10644599. Copyright may be transferred without notice, after which this version may no longer be accessible. }%
}

\begin{document}

\maketitle
\thispagestyle{empty}
\pagestyle{empty}

\begin{abstract}

This paper presents a novel fleet management strategy for battery-powered robot fleets tasked with intra-factory logistics in an autonomous manufacturing facility. 
In this environment, repetitive material handling operations are subject to real-world uncertainties such as blocked passages, and equipment or robot malfunctions.
In such cases, centralized approaches enhance resilience by immediately adjusting the task allocation between the robots.
To overcome the computational expense, a two-step methodology is proposed where the nominal problem is solved a priori using a Monte Carlo Tree Search algorithm for task allocation, resulting in a nominal search tree.
When a disruption occurs, the nominal search tree is rapidly updated a posteriori with costs to the new problem while simultaneously generating feasible solutions.
Computational experiments prove the real-time capability of the proposed algorithm for various scenarios and compare it with the case where the search tree is not used and the decentralized approach that does not attempt task reassignment.

\end{abstract}

\section{INTRODUCTION}\label{sec:Intro}

Advancements in material handling have identified fleets of autonomous mobile robots and automated guided vehicles as key components of autonomous operations within flexible manufacturing systems (FMS) \cite{Flexible}.
These battery-powered robots handle the repetitive material handling tasks (MHT) integral to manufacturing a product \cite{fragapane2022increasing}.
In this context, an energy-aware fleet management policy aims to assign tasks and route robots in a manner that minimizes energy expenses while adhering to the constraints imposed by battery charging policies \cite{scholz2023decentral}.
Furthermore, the operational requirements of an FMS introduce additional constraints that ensure that routes visit pick up locations before the corresponding delivery locations, and also respect the robot's payload limits.

This paper examines the scenario where a robot fleet, while actively executing a nominal MHT based on a pre-determined fleet policy, experiences disruptions due to real-world uncertainties.
These perturbations in the definition of the MHT arise from various factors such as machine failures, robot malfunctions, battery degradation, fluctuating charge power or blocked passageways due to fallen objects \cite{daub2020optimizing}.
These disruptions can drastically affect the optimal solution and the nominal policy may no longer be optimal or even feasible.
Resilience in this context refers to the ability of the fleet management system to immediately adapt the fleet policy to guarantee uninterrupted operations \cite{beyer2007robust}.

The $\mathcal{NP}$-hard nature of the task assignment problem has resulted in the development of numerous decentralized approaches for fleet management that focus on reassigning tasks only to the affected robots to rapidly recover from a perturbation \cite{sauer2022decentralized}.
However, this approach typically leads to suboptimal solutions as it overlooks the possible assistance that other unaffected AMRs in the fleet could provide to enhance resilience \cite{fragapane2021planning}.
In contrast, centralized fleet management (CFM) considers all the available AMRs for task reassignment and rerouting, thereby harnessing the fleet's collective resilience to achieve an optimal policy \cite{fragapane2021planning}.
However, using exact methods for CFM is intractable for real-time execution due to the $\mathcal{NP}$-hardness of the problem \cite{de2020automated}.
For this reason, metaheuristic algorithms such as simulated annealing, genetic algorithms, and tabu search are typically used to quickly improve multiple trial solutions \cite{rios2021recent, Elshaer2020AVariants}.
However, these methods do not provide algorithmic guarantees on the convergence or optimality of the resulting solution \cite{fahimnia2018planning}.
Another approach that enables real-time computation uses supervised machine learning to map expert-identified problem perturbations to pre-computed solutions \cite{malus2020real, lodi2020learning}.
However, its performance is adversely affected when the disruption differs significantly from the training data.

A gap in the CFM literature in the context of repetitive MHTs is the under-utilization of prior knowledge of the nominal search space when recomputing the policy for a perturbed problem.
This stems from the intractable memory requirements needed to store information about the task assignment and routing search space when using the typical approach of using a single decision variable to define both the task assignment and the robot routes\cite{fachini2020logic}.
Consequently, an entirely new problem is solved each time a small change to the nominal problem is realized, restricting real-time applicability to small problems \cite{de2020automated}.

In this paper, the routing problem is solved using a heuristic and the task assignment search space is explored using a Monte Carlo Tree Search (MCTS) algorithm.
For task assignment problems, MCTS algorithms build a search tree that stores cost estimates of task assigning decisions while exploring the search space and producing solutions with low optimality gaps \cite{Edelkamp2015Monte-CarloLogistics, Barletta2022HybridSearch}.
The contribution of this paper lies in the re-utilization of the search tree topology and cost estimates as prior knowledge when the problem is perturbed.
This prior knowledge is used in a transfer learning framework to rapidly update cost estimates while also generating solutions, after which the MCTS algorithm is re-initialized.
Computational experiments are performed on a modified TSPLIB instance \cite{reinhelt2014tsplib} to capture realistic FMS operational constraints of charging policies, payload and battery constraints, and pickup - delivery requirements of unique items.
Results show that the solutions obtained using prior knowledge have a lower optimality gap than when the perturbed problem is approached as an entirely new problem.


\section{PROBLEM DEFINITION}\label{Sec:ProbForm}

Consider that $n$ material handling tasks are to be completed, and the different commodities are represented by the set $\mathcal{H}:=\{h_1,h_2,...,h_n\}$.
The set of paired pickup and delivery locations are defined by $\mathcal{V}^P:=\{1,2,...,n\}$ and $\mathcal{V}^D:=\{n+1,n+2,...,2n\}$ respectively.
Define $\mathcal{V}:=\mathcal{V}^P\cup \mathcal{V}^D$, and let each location $i\in\mathcal{V}$ be associated with a cargo mass $q_{im}\in\mathbb{R}, \forall m\in\mathcal{H}$.
A commodity picked up at $i\in\mathcal{V}^P$ is paired with a delivery location $n+i \in \mathcal{V}^D$, such that $q_{im}+q_{i+n,m}=0$.
The start and end locations of each robot are at the depot defined by the nodes $\{0,2n+1\}$, and also serve as the charger location.
Define $\overline{\mathcal{V}}:=\mathcal{V}\cup \{0,2n+1\}$ so that the graph representation is given by $\mathcal{G}:=(\overline{\mathcal{V}},\mathcal{E})$, where $\mathcal{E}:=\{(i,j)\in \overline{\mathcal{V}}\times\overline{\mathcal{V}}: i\neq j\}$ denotes the set of edges.
The set $\mathcal{T} := \{1,2,...,t_{max} \}$ denotes the types of robots available for performing the defined material handling tasks.
For each type of robot $t\in \mathcal{T}$, the battery size is defined by $B^t$ and its payload capacity is $Q^t$.
The set $\mathcal{R}^t := \{1, 2, ..., r^t_{max} \}$ represents robots of type $t$ present in the fleet.
For each type $t\in\mathcal{T}$, the energy to travel between each node pair $(i,j)\in \overline{\mathcal{V}}$ is defined as parameter $\delta e_{ij}^t\in \mathbb{R}^+$, normalized to be a fraction of the battery capacity $B^t$. 
The problem formulation is defined in Eq.\ref{eq:ProbForm} below:

\begin{subequations}  \label{eq:ProbForm}
\allowdisplaybreaks
\begin{align}  
    \label{eq:obj}      &J  = \min_{ \substack{x_{ij}^{a_t}\\ }}  ~\sum_{(ij) \in \mathcal{E} } ~ \sum_{t\in \mathcal{T}} ~ \sum_{a_t\in \mathcal{R}^t} E_{ij}^{a_t} x_{ij}^{a_t}\\
    \label{eq:binary}   \textrm{s.t.} \quad &x_{ij}^{a_t}\in \{0,1\}       \quad \quad \forall (i,j)\in \mathcal{E}, ~   a_t\in \mathcal{R}^t, ~ t\in \mathcal{T} \\
    \label{eq:depot+}   &\sum_{j\in \mathcal{V}^p} x_{0j}^{a_t} \leq 1    ~~ \quad  \forall   a_t\in \mathcal{R}^t, ~ t\in \mathcal{T} \\
    \label{eq:depot-}   &\sum_{i\in \mathcal{V}^D} x_{i,2n+1}^{a_t}  \leq 1            \quad \forall   a_t\in \mathcal{R}^t, ~ t\in \mathcal{T} \\ 
    \label{eq:once}   &\sum_{(ij) \in \mathcal{E}} x_{ij}^{a_t}  \leq 1            ~\quad\quad\quad \forall   a_t\in \mathcal{R}^t, ~ t\in \mathcal{T} \\ 
    \label{eq:through}  &\sum_{i\in \mathcal{V}} x_{ij}^{a_t} = \sum_{k \in \mathcal{V}} x_{jk}^{a_t}       \quad \forall j \in \mathcal{V}, ~ a_t\in \mathcal{R}^t,~  t\in \mathcal{T}\\
    \label{eq:cargo init}       &y_{0m}^{a_t} = 0                                                          \quad \forall m\in\mathcal{H}, a_t\in \mathcal{R}^t,~  t\in \mathcal{T}\\
\begin{split} \label{eq: mass evolution}
                                &y_{jm}^{a_t} = y_{im}^{a_t} + \sum_{i\in \mathcal{V}}q_{jm} x_{ij}^{a_t}  \\
                                & \quad \quad \quad \quad \quad \forall m\in\mathcal{H}, j \in \mathcal{V}, ~ a_t\in \mathcal{R}^t,~  t\in \mathcal{T}
\end{split}\\    
\begin{split} \label{eq:cargoPrecedence}
                                &\sum_{i\in \mathcal{V}}y_{im}^{a_t} x_{ij}^{a_t} = -q_{jm}  \\
                                &\quad \quad \quad \quad \quad  \forall m\in\mathcal{H},j \in \mathcal{V}^D, a_t\in \mathcal{R}^t,~  t\in \mathcal{T}
\end{split}\\ 
\label{eq:cargo limit}      &\sum_{m\in \mathcal{H}}y_{im}^{a_t} \leq Q^t                           \quad \quad \forall i \in \overline{\mathcal{V}}, a_t\in \mathcal{R}^t,~  t\in \mathcal{T}\\
\label{eq:socEvolution}      &z_j^{a_t} = \begin{cases}
        z_i^{a_t}-\delta e_{ij}^t    &\text{if } x_{ij}^{a_t} =1 \wedge z_i^{a_t}-\delta e_{ij}^t>0\\
        1-\delta e_{0j}^t            &\text{if } x_{ij}^{a_t} =1 \wedge z_i^{a_t}-\delta e_{ij}^t\leq 0\\
    \end{cases}\nonumber \\
    &\quad  \quad \quad \quad \quad~~ \quad \quad \forall (i,j)\in \mathcal{E}, ~   a_t\in \mathcal{R}^t, ~ t\in \mathcal{T}\\
\label{eq:battery} & z_0^{a_t} =1;~0 \leq z_i^{a_t} \leq 1 ~~ \forall i\in \overline{\mathcal{V}}, ~   a_t\in \mathcal{R}^t, ~ t\in \mathcal{T} \\
\label{eq:energyEvolution}      E_{ij}^{a_t} &= \begin{cases}
        B^t\delta e_{ij}^t             &\text{if } x_{ij}^{a_t} =1 \wedge z_i^{a_t}-\delta e_{ij}^t>0\\
        B^t(\delta e_{i0}^t+\delta e_{0j}^t)           &\text{if } x_{ij}^{a_t} =1 \wedge z_i^{a_t}-\delta e_{ij}^t\leq 0\\
    \end{cases}\nonumber \\
    &\quad  \quad \quad \quad \quad~~ \quad \quad \forall (i,j)\in \mathcal{E}, ~   a_t\in \mathcal{R}^t, ~ t\in \mathcal{T}
        \end{align} 
\end{subequations}
The goal of minimizing the total energy traveled by all robots in the fleet is captured in Eq. (\ref{eq:obj}) of the MHT problem formulation.
Here, $E_{ij}^{a_t} \in\mathbb{R}^+$ accounts for charge events and is the energy expense of a robot $a_t\in \mathcal{R}^t$ between a pair of nodes $(i,j)\in \mathcal{E}$.
Binary variables $x_{ij}^{a_t}$ are used to indicate whether robot $a_t$ of type $t\in \mathcal{T}$ uses edge $(i,j)\in \mathcal{E}$.
If a robot is assigned a task, it must start and end at the depot, as specified by Eq. (\ref{eq:depot+}) and (\ref{eq:depot-}) respectively. 
Additionally, the robot is permitted to visit each location at most once, as enforced by Eq. (\ref{eq:once}), and must leave the location after completing the visit, as defined in Eq. (\ref{eq:through}).
Payload variables $y_{im}^{a_t}$ are used to define the mass of commodity $m\in\mathcal{H}$ being carried by robot $a_t\in\mathcal{R}^t$ as it leaves node $i\in \overline{\mathcal{V}}$.
All robots start their tour with no payload at the depot, as defined in Eq. (\ref{eq:cargo init}).
The evolution of the commodity-wise payload is defined in Eq. (\ref{eq: mass evolution}) as the robot visits locations in its tour.
Precedence constraints for each commodity are defined in Eq. (\ref{eq:cargoPrecedence}), meaning that a robot can visit a delivery location if and only if the corresponding commodity has been picked up previously.
Payload limitations are captured in Eq. (\ref{eq:cargo limit}).
The state of charge (SOC) of robot $a_t$ as it arrives at location $j$ is given by $z_j^{a_t}$ and Eq. (\ref{eq:socEvolution}) defines the charging policy that requires a robot to head to the depot for a recharge if required.
As described in Eq. (\ref{eq:energyEvolution}), the energy expense $E_{ij}^{a_t}$ between locations $i$ and $j$ is dependent on whether a recharge event occurs between the two locations.
\begin{figure}[t]
    \centering
    \vspace{0mm}
    \includegraphics[trim =0mm 79mm 155mm 0mm, clip, width=0.58\linewidth]{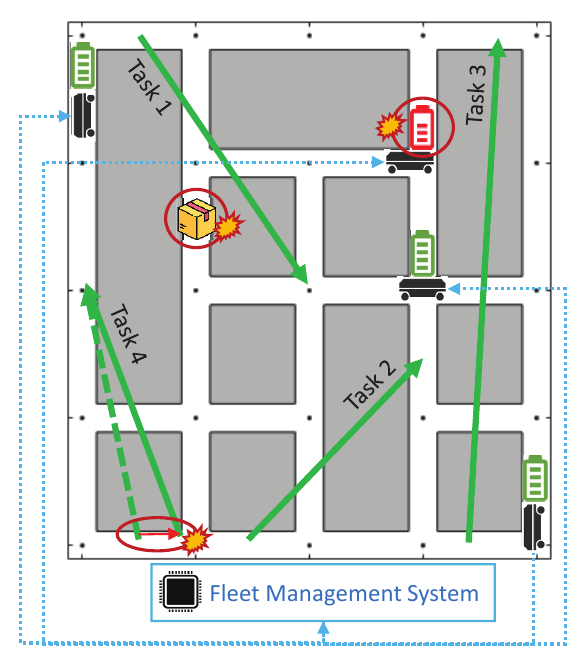}
    \caption{Illustration of perturbations to the problem formulation}
    \label{Im:Illustrative example}
    \vspace{-5mm}
\end{figure}

The uncertainties of real-world deployment cause disruptions that manifest as changes in the parameters of Eq. (\ref{eq:ProbForm}).
In Fig. \ref{Im:Illustrative example}, an illustrative plant layout is shown, with green arrows marking 4 material handling tasks and blue dotted lines indicating the centralized communication between the FMS and the robots.
A blocked aisle or a change in pickup position results in a change in parameter $\delta e_{ij}$ in Eq. (\ref{eq:ProbForm}).
Similarly, a degraded robot battery capacity changes parameter $B^t$.
The FMS objective is to quickly reassign tasks and routes to the robots to adapt to the updated MHT definition.

\section{METHODOLOGY}\label{Sec:Method}
The fleet policy both assigns tasks to the robots and also routes each robot, and is defined by the value of the binary variable $x$.
Finding the optimal solution to Eq. (\ref{eq:ProbForm}) is computationally expensive because of the $\mathcal{NP}$-hardness of the problem and the nonlinear constraints.
The proposed framework uses an offline MCTS algorithm to first compute a near-optimal solution to the nominal problem.
This utilizes a sufficiently high computational time budget since the MHT parameters are known well in advance.
This produces a richly populated MCTS search tree with cost estimates for task assignment decisions.
When a perturbation is realized, an online algorithm uses these cost estimates to rapidly obtain feasible solutions to the updated problem.

\subsection{Solving the nominal problem offline} \label{Sec:MCTSoffline}
\begin{figure}[t]
    \centering
    \includegraphics[trim =0mm 78mm 80mm -2mm, clip, width=0.98\linewidth]{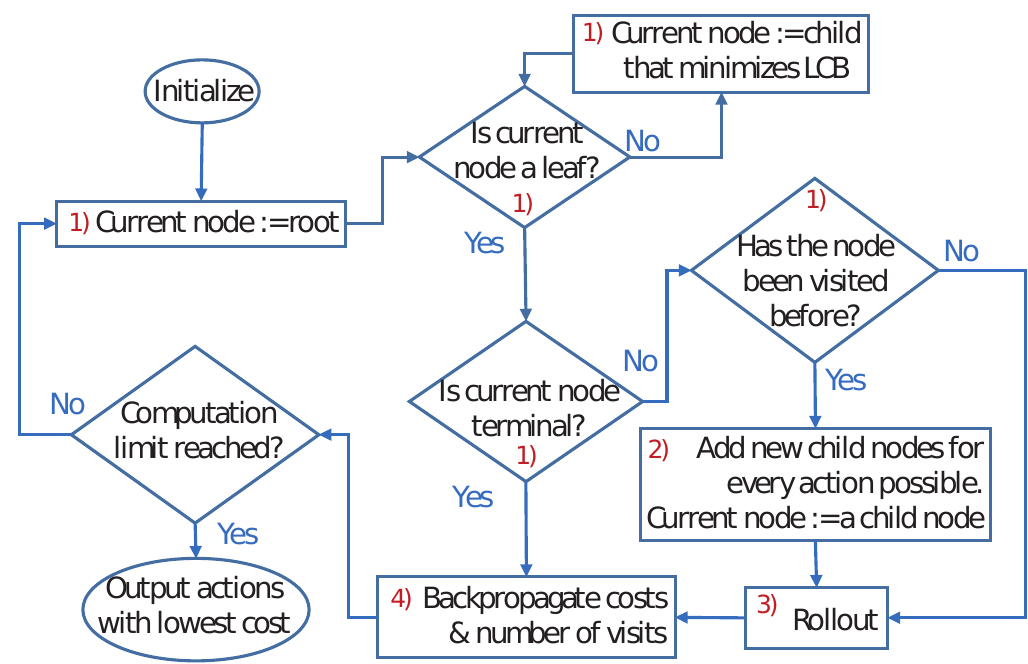}
    \caption{Flow diagram of the Monte Carlo Tree Search Algorithm}
    \label{Im:UCT-MH-Schematic}
    \vspace{-6mm}
\end{figure}

The MCTS algorithm explores the task assignment search space, generating a search tree whose nodes represent decisions related to assigning a robot to a task.
The root node represents the start of the decision-making process where no tasks have been assigned.
The terminal node of the tree represents the final outcome of the task-assigning process, signifying that all the tasks have been assigned to the available robots.
As the tree is traversed from the root node to a terminal node, tasks are assigned to robots based on their order in the defined task list.
The parent of a node is the node that precedes it in the decision-making process, that is, the robot assigned the previous task in the task list.
Similarly, its child nodes are the nodes that immediately follow it, and represent the robots available for selection at the next task in the task list.
Each child node is connected to its parent by a branch that represents the decision of assigning the next task to that robot while fixing the previous decisions from the parent node to the root node.
A leaf node does not have any child nodes, and if non-terminal, indicates that some task-assignment decisions have not yet been made.
The MCTS algorithm is as enumerated in Fig. \ref{Im:UCT-MH-Schematic}:

\textbf{1) Selection:} Starting from the root node, the algorithm traverses the search tree by selecting child nodes based on a selection policy.
For the cost minimization objective, the Lower Confidence Bound (LCB) selection policy is used:
        \begin{equation}
            \text{LCB}(s) = \operatorname*{arg\,min}_{s' \in \text{children of } s} \frac{J(s')}{N(s')J_{max}} - \gamma \sqrt{\frac{ \ln N(s)}{N(s')}}
            \label{eq:LCB}
        \end{equation}
where $N(s)$ is the number of cost explorations at node $s \in S$ and $S$ is the set of nodes that constitute the search tree.
$J(s)$ is the sum of costs from all the previous visits to node $s$.
The constant $\gamma$ balances the exploitation of promising nodes with the exploration of unfavorable nodes that are visited less often.
This ensures that the entire search space is systematically explored when given sufficient computation time.
During the conducted explorations, the maximum cost found is denoted by $J_{max}$, and is used as a normalization factor that is continually updated as the search proceeds.

The process of selecting child nodes by applying Eq. (\ref{eq:LCB}) starts from the root node and continues until a leaf or terminal node is reached.
If the selected node is a previously visited child node, an expansion of the tree is conducted as defined in \textit{step 2)} while if it was unvisited, a rollout is conducted per \textit{step 3)}. 
On the other hand, if it is a terminal node, the costs of routing are computed for each robot and the costs are backpropagated according to \textit{step 4)}.

\textbf{2) Expansion:} 
After selecting a node based on the LCB policy, the tree is expanded by adding child nodes to the selected node to represent possible task assignments.
Once new child nodes have been added, one is selected for rollout.

\textbf{3) Rollout:} If the selected node is a leaf node, Monte Carlo sampling randomly assigns the remaining tasks to fleet robots until the terminal node is reached.
Here, a robot has been assigned to each task and the total cost associated with the assignment is obtained by solving the routing problem for each robot.
While numerous approaches exist to obtain the routing cost for a single robot, in this paper the recursive B\&B Alg. \ref{Alg:TSP} derived from \cite{baltussen2023parallel} is used to account for the nonlinear constraints associated with precedence and charging policies.
To limit computation time, the B\&B of Alg. \ref{Alg:TSP} is terminated after a 0.1 second time cap, since reasonably good routes are expected due to the best first order of exploration in the recursive algorithm.
The total routing cost for the entire fleet then provides an estimate of the cost of selecting that node in \textit{Step 1}.

\textbf{4) Backpropagation:} 
To update the tree based on the outcome of the conducted rollouts, the algorithm traverses the search tree from the selected leaf or terminal node $s_l$ up to the root node. 
For each parent node $s$ whose selection by Eq. (\ref{eq:LCB}) resulted in the evaluation at $s_l$, the number of visits is updated as $N(s) \leftarrow N(s) + r$, where $r$ is the number of rollouts conducted.
The accumulated costs for these nodes are also updated as $J(s) \leftarrow J(s) + \sum_{i=1}^{r} J_r(s_l)$, where $J_r(s_l)$ represents costs obtained from the rollout at $s_l$.

The four steps are repeated until the pre-defined and problem-specific computational budget of time or number of iterations is exhausted.
Throughout the MCTS exploration, the task assignment that resulted in the minimum cost $J_{min}$ is referred to as its incumbent solution.
Like $J_{max}$, the incumbent solution is also continually updated as the search progresses and is the output of the MCTS algorithm when terminated.
The resulting search tree topology and the average cost $J(s)/N(s)$ at each node $s$ are saved as the prior knowledge of the nominal problem which will be utilized when a perturbation occurs.
\begin{algorithm}
      \centering
      \footnotesize
      \caption{Routing B\&B}\label{Alg:TSP}
            \begin{algorithmic}[1]
            \STATE sequenceCost  = \textbf{\emph{B\&B}(robotState, taskList, \emph{location})}
            \STATE Find feasible next locations based on payload, cargo, SOC
            \STATE Sort locations by operational cost of branching to that location 
            \FOR{\emph{i} in feasible locations}
                \STATE \emph{branchCost} = tourCost + operational cost(i)
                \IF{\emph{branchCost} $\geq$ robotState.bestCost}
                    \STATE continue \{ skip to next \emph{location i+}\}
                \ELSIF{\emph{branchCost}$<$ robotState.bestCost}
                    \STATE \emph{State+} = Update robotState: SOC, position, remaining locations
                        \IF{number of remaining locations $> 0$}
                            \STATE Cost = \textbf{\emph{B\&B}(robotState+, taskList, \emph{location}(i))} \text{Recursive Alg. \ref{Alg:TSP}}
                        \ELSE
                            \STATE State.bestCost = Cost
                        \ENDIF
                \ENDIF
            \ENDFOR
            \STATE \textbf{Return} robotState
            \end{algorithmic}
      \end{algorithm}
\subsection{Solving the perturbed problem online} \label{recomputation}
\begin{figure}[b]
    \centering
    \vspace{-3mm}
    \includegraphics[trim = 0mm 90mm 65mm 1mm, clip, width=0.99\linewidth]{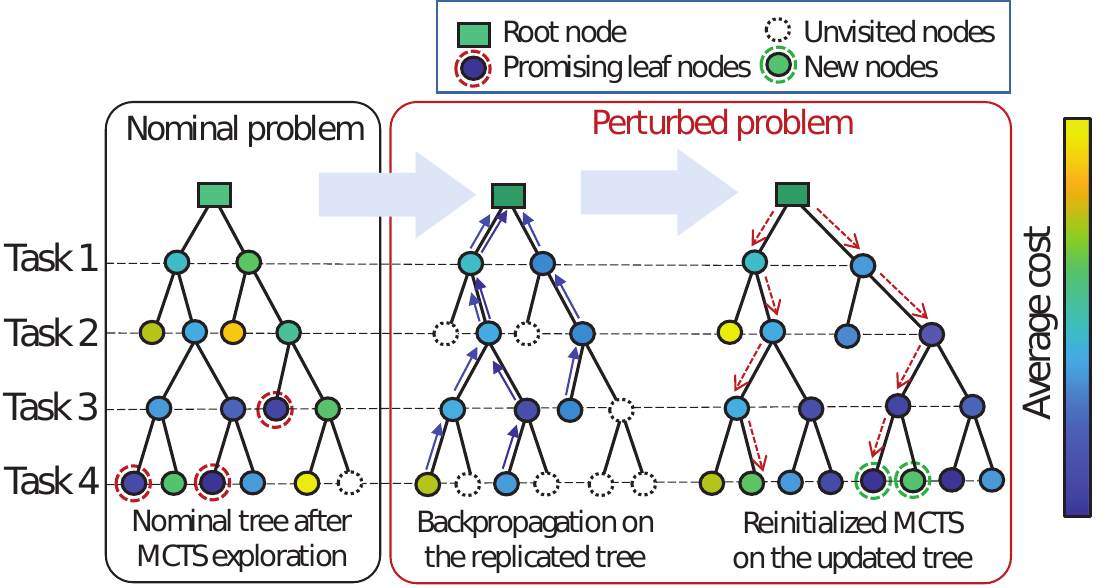}
    \caption{Schematic of the proposed algorithm}
    \label{Im:re-exploration}
\end{figure}
Consider the case where the nominal Eq (\ref{eq:ProbForm}) has been addressed using the offline MCTS algorithm, generating a task assignment search tree, referred to as the nominal tree.
In real-world operation, when perturbations affect the problem definition, the proposed method makes use of the nominal tree topology and cost information as follows:
%

\begin{enumerate}
    \item The leaf nodes $s_l$ of the nominal search tree are first ordered in increasing average costs to the nominal problem given by $J(s_l)/N(s_l)$.
    \item A search tree that replicates the topology of nodes and branches in the nominal tree is initialized. However, at each node of this perturbed search tree, the number of visits $N(s)$ and accumulated costs $J(s)$ are set to zero.
    \item For a predefined parameter $k$, select the $k^{th}$ percentile of ordered leaf nodes. Rollouts are then conducted for these nodes in a nominally cheapest-first order.
    During the rollout process, leaf node costs are re-evaluated under the perturbed problem parameters. 
    The updated costs and number of visits are then backpropagated through the perturbed tree as defined in \textit{step 4)} of Section \ref{Sec:MCTSoffline}.
    This simultaneously updates the search tree while exploring promising leaf nodes.
    The incumbent solution to the perturbed problem is continuously updated and available for a policy update if necessary.
    \item Once all the selected leaf nodes have undergone re-evaluation, the MCTS algorithm, as described in Section \ref{Sec:MCTSoffline}, is re-initialized on the updated perturbed tree. 
    This creates new nodes and utilizes the balance of exploitation and exploration to further reduce the cost for the duration of the remaining computation time.
\end{enumerate}

The proposed method is illustrated in Fig. \ref{Im:re-exploration}, demonstrating how the topology of the nominal search tree is replicated, and the costs associated with promising leaf nodes are re-evaluated with backpropagation.
The figure also shows the generation of new nodes within the perturbed tree once the online MCTS algorithm is initialized in Step 4). 
An inherent assumption in this approach is that the perturbation does not change the topology of the existing search tree, that is, the number of robots is not changed, since this would not permit the topology of the nominal tree to be reused.
When perturbations are bounded, which is a reasonable assumption for the controlled environment of an FMS, the online approach is expected to yield solutions with a reduced optimality gap compared to not utilizing the nominal tree.

\section{Computational Experiments}

To test the effectiveness of the proposed algorithm across a variety of perturbation types, the nominal MHT problem was first defined using a TSPLIB benchmark instance \cite{reinhelt2014tsplib}: 
\begin{enumerate}[\textit{Step} 1:]
    \item Load the TSPLIB $eil51$ point cloud to obtain a set of $n$ points with defined Cartesian coordinates.
    \item Find the centroid of the point cloud.
    \item Sort and assign indices $1,2,...,n$ to the points by order of increasing distance from the centroid.
    \item Designate the point with index 1 as the depot.
    \item Define precedence constraints between points with pairs of indices as $(2\prec n),(3\prec n-1)$, and so on.
\end{enumerate}

The resulting point cloud defines a depot and 25 MHTs, each with paired pickup and delivery locations.
In the nominal case being studied, the MHTs are to be completed by two robots with payload capacities $Q^1=Q^2=10$ commodities.
The energy expense associated with traveling a Euclidean distance of $d$ units is defined to be $d$ kJ, and the battery capacity of each robot is nominally $B^1 = B^2 = 20$ kJ.

Extensive computational experiments were conducted in a Matlab R2022a environment on an Intel Xeon E5-2680 v4 CPU clocked at 2.4 GHz at the Ohio Super Computer \cite{OSC}.
Three fleet management strategies were compared for perturbations associated with battery degradation, payload capacity variations, and shifts in pickup and delivery locations.
The optimal task assignment solution was first found for the nominal problem and each perturbation, requiring 370 hours of processing time each.
This involved an exhaustive search that first listed the $2^{25}$ possible task assignments, and then obtained the cost of each task assignment using Alg. \ref{Alg:TSP}, after which the lowest-cost task assignment was found.

Each experiment was repeated 25 times to account for the stochastic nature of MCTS algorithms.
In each repetition, the nominal problem was first solved using the offline algorithm of Section \ref{Sec:MCTSoffline} to populate a nominal search tree over a computational time budget of 12 hours.
The 25 resulting task assignment solutions varied slightly due to the stochasticity but were within 5\% of the optimal solution.

For benchmarking, a decentralized approach was evaluated that continued to use one of the 25 nominal task assignment solutions that were computed offline.
The problem perturbation was only addressed by rerouting the affected robots by using Alg. \ref{Alg:TSP}, providing near-instantaneous recovery from the perturbation.
Another comparison was made with a centralized approach which executed the offline algorithm of Section \ref{Sec:MCTSoffline} ab initio when a perturbation is realized, without utilizing nominal search tree information.
This was also repeated 25 times to account for stochasticity.
Finally, to evaluate the proposed online CFM algorithm, each of the 25 nominal trees was used as prior knowledge, thus producing 25 updated search trees for each perturbation.

For every MCTS algorithm used in these experiments, the parameter $\gamma$ in Eq. (\ref{eq:LCB}) is set to $\sqrt{0.5}$, and the number of rollouts $r$ is set to 20.
When the nominal tree is utilized, the parameter $k$ of the online algorithm is set to 0.05, implying that the 5th percentile of low-cost nominal leaf nodes is first explored to acquire updated costs for the new problem.


\begin{figure}[b]
    \centering
    \vspace{-4.5mm}
     \subfloat[Incumbent solutions obtained by the algorithms \label{Im: B16_scatter}]{%
        \includegraphics[trim =10mm 0mm 12mm 3.3mm, clip, width=0.97\linewidth]{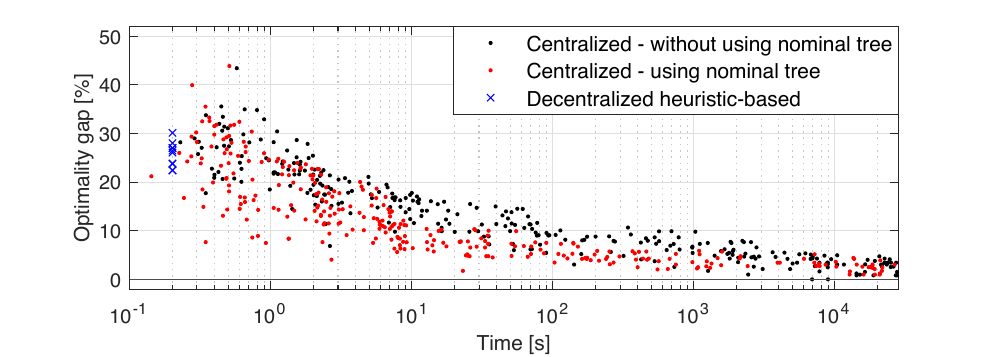}}
\\
\vspace{-3mm}
      \subfloat[1 second \label{Im: B16_histo_1s}]{%
        \includegraphics[trim =0mm 0mm 0mm 0mm, clip, width=0.31\linewidth]{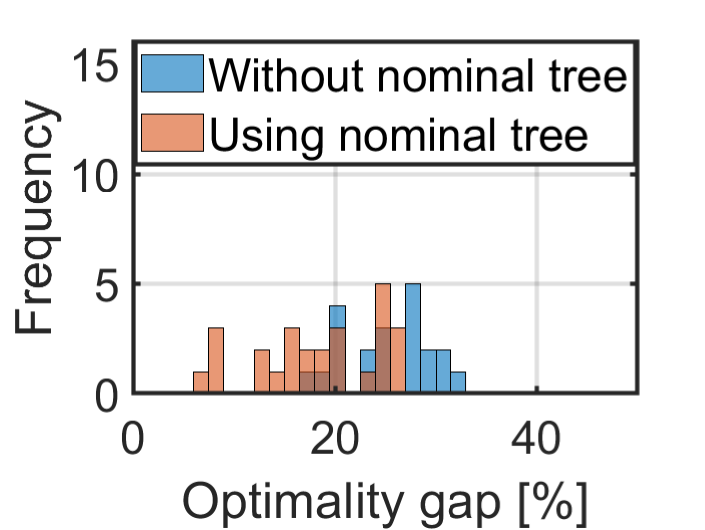}}
    \hspace{.02\linewidth}
  \subfloat[1 minute \label{Im: B16_histo_1m}]{%
        \includegraphics[trim =0mm 0mm 0mm 0mm, clip, width=0.31\linewidth]{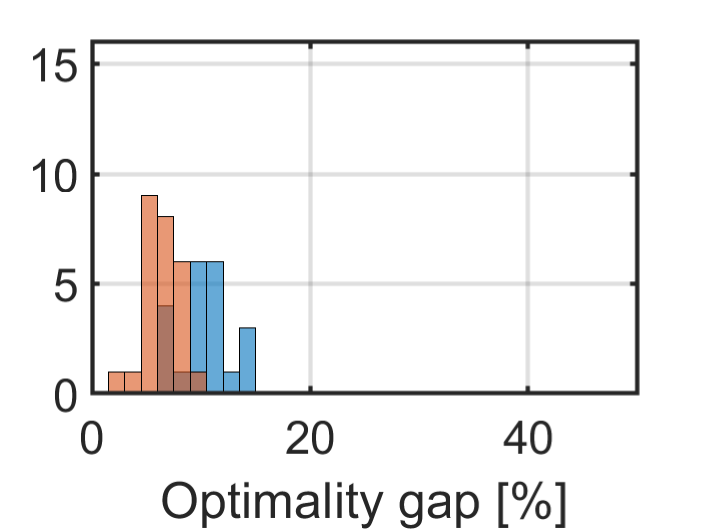}}
    \hspace{.02\linewidth}
  \subfloat[1 hour\label{Im: B16_histo_1h}]{%
        \includegraphics[trim =0mm 0mm 0mm 0mm, clip, width=0.31\linewidth]{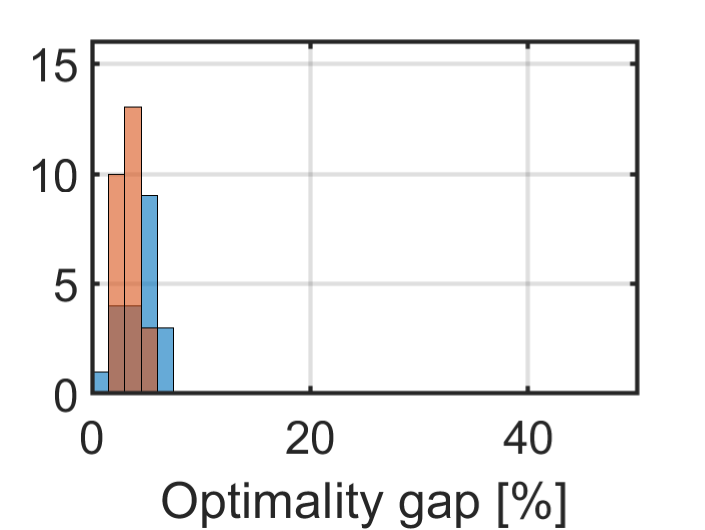}}
  \caption{Battery capacity of one robot changed from 20 to 16 kJ}
  \vspace{0mm}
  \label{Im: batt16} 
\end{figure}

\subsection{Variation in battery capacity}
\begin{figure}[t]
    \centering
    \vspace{1.5mm}
     \subfloat[Incumbent solutions obtained by the algorithms \label{Im: B12_scatter}]{%
        \includegraphics[trim =10mm 0mm 12mm 1.6mm, clip, width=0.97\linewidth]{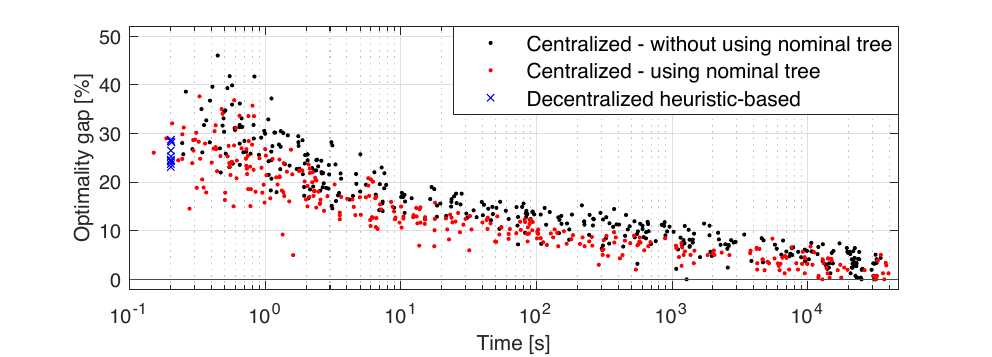}}
\\
\vspace{-3mm}
      \subfloat[1 second \label{Im: B12_histo_1s}]{%
        \includegraphics[trim =0mm 0mm 0mm 0mm, clip, width=0.31\linewidth]{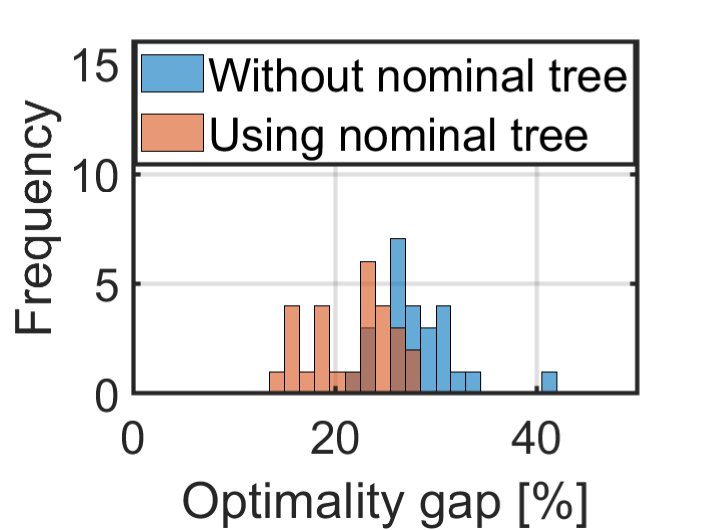}}
    \hspace{.02\linewidth}
  \subfloat[1 minute \label{Im: B12_histo_1min}]{%
        \includegraphics[trim =0mm 0mm 0mm 0mm, clip, width=0.31\linewidth]{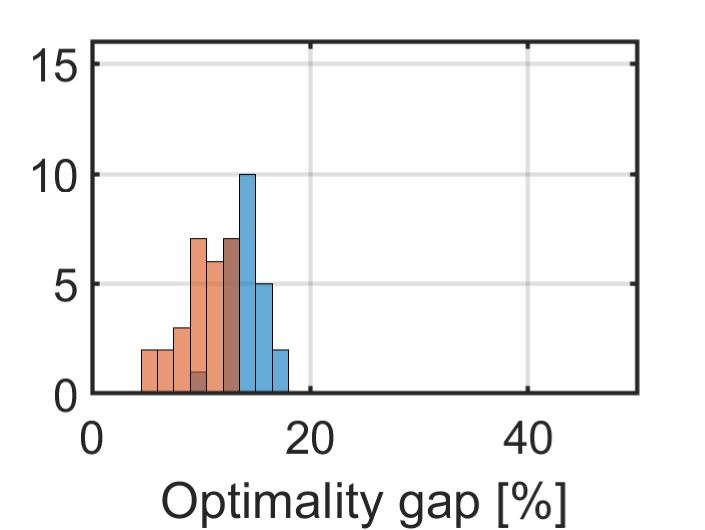}}
    \hspace{.02\linewidth}
  \subfloat[1 hour\label{Im: Im: B12_histo_1h}]{%
        \includegraphics[trim =0mm 0mm 0mm 0mm, clip, width=0.31\linewidth]{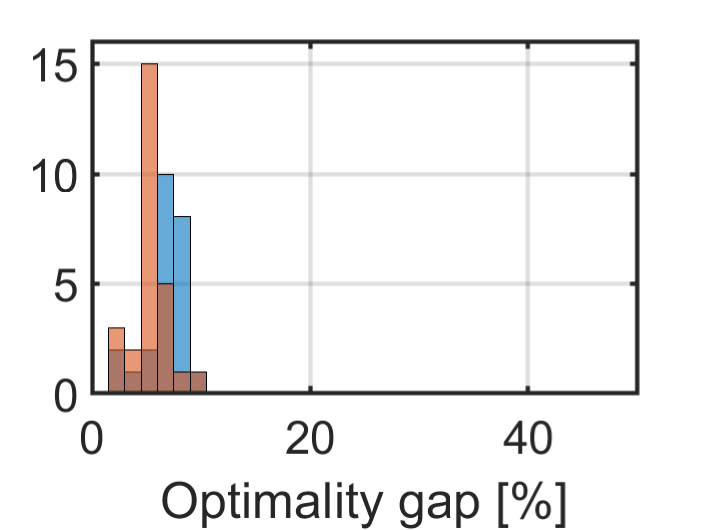}}
  \caption{Battery capacity of one robot changed from 20 to 12 kJ}
  \vspace{-5mm}
  \label{Im: batt12} 
\end{figure}

The battery capacity of one of the robots was perturbed from its original capacity of $20$ kJ to $16$ kJ, without altering the capacity of the other robot.
Since the decentralized method does not optimize the task assignment, the 25 task assignments obtained from the repeated solving of the nominal problem produced 25 heuristic solutions, many of which overlap, as seen in Fig. \ref{Im: B16_scatter}.
These solutions are found near-instantaneously because only the time-capped heuristic Alg. \ref{Alg:TSP} is used, but they are significantly outperformed within 10 seconds by every repetition of the centralized methods.
In the case of these centralized algorithms, it is evident that when the perturbed problem is solved without utilizing the nominal tree, the incumbent solution costs at any instant are typically higher than when the nominal tree is utilized.
Histograms of incumbent solutions at one second, minute, and hour of computation time are shown in Fig. \ref{Im: B16_histo_1s}, \ref{Im: B16_histo_1m}, and \ref{Im: B16_histo_1h} respectively, indicating a significant advantage to using the nominal tree, especially when computation time is limited.
Given sufficient computation time, it is seen that both the centralized algorithms converge to the optimum solution in each of their 25 repetitions.
Similar results are seen in Fig. \ref{Im: batt12} for the case when $B^2$ is changed to $12$ kJ.


\begin{figure}[b]
    \centering
    \vspace{-11mm}
      \subfloat[Nominal problem($\xi = 0\%$) \label{Im: nominal def}]{%
        \includegraphics[trim =0mm 0mm 0mm 0mm, clip, width=0.48\linewidth]{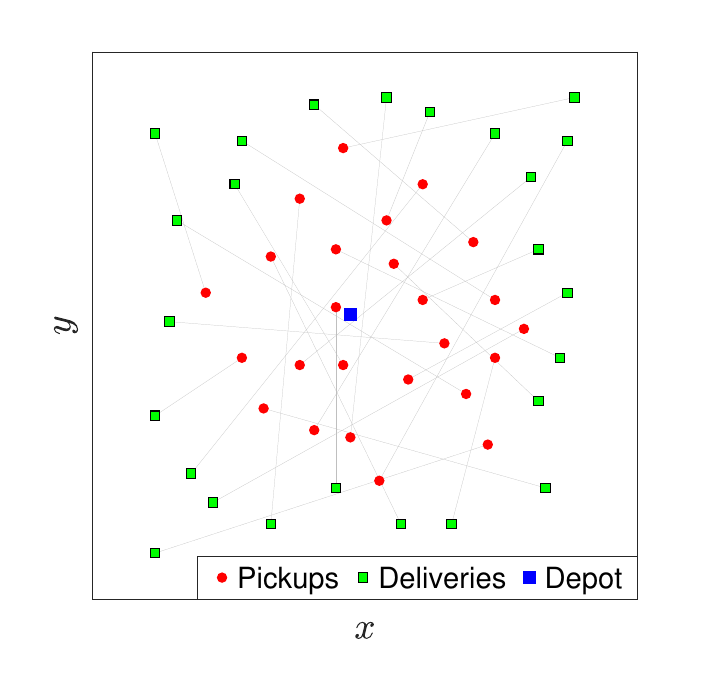}}
    \hspace{.02\linewidth}
  \subfloat[Perturbed problem: $\xi = 4\%$ \label{Im: box of 10}]{%
        \includegraphics[trim =0mm 0mm 0mm 0mm, clip, width=0.48\linewidth]{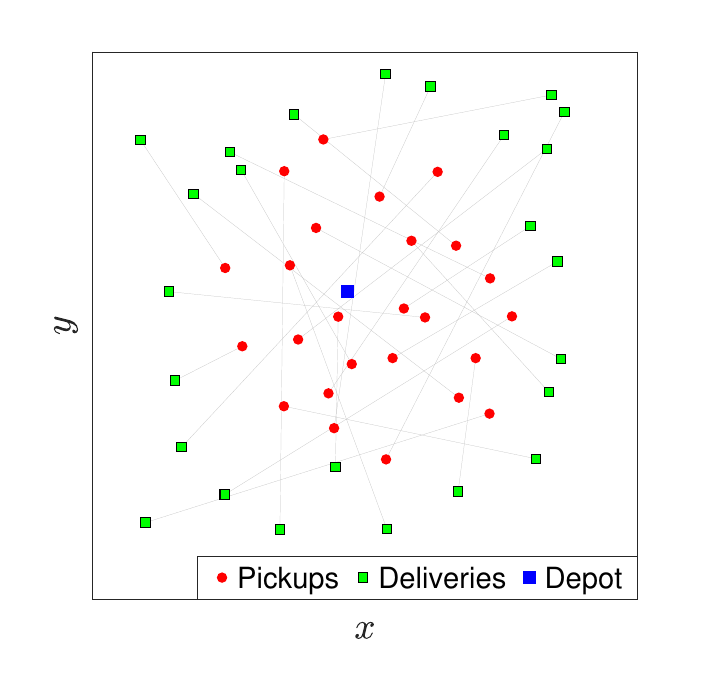}}
               \\
  \subfloat[Incumbent solutions obtained for $\xi$ = 4\% \label{Im: ResultsLoc4}]{%
        \includegraphics[trim =10mm 0mm 12mm 3.3mm, clip, width=0.97\linewidth]{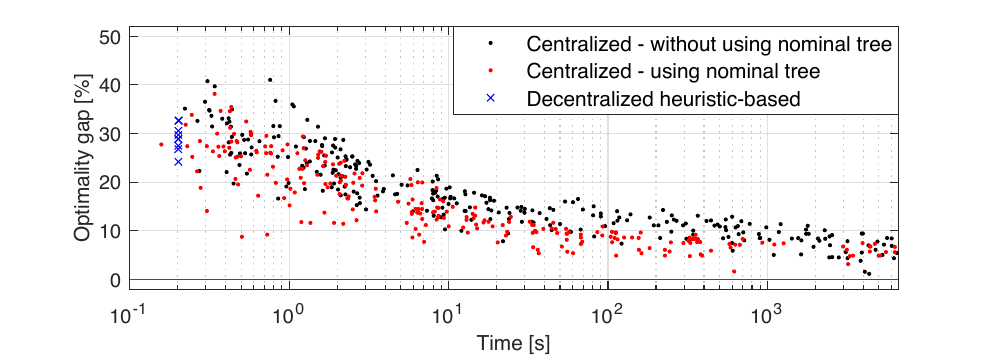}}
  \caption{Spatial variations in pickup and delivery locations}
  \vspace{0mm}
  \label{Im: spatial Perturb} 
\end{figure}

\subsection{Spatial variations of locations}


The $eil51$ point cloud is shown in Fig. \ref{Im: nominal def}, where grey line segments show the precedence constraints.
Let $\overline{x}$ and and $\overline{y}$ denote the range of $x$ and $y$ coordinates respectively.
The box uncertainty parameter, denoted by $\xi$, causes deviations within the range of $\pm \xi\overline{x}$ and $\pm \xi\overline{y}$ in the $x$ and $y$ coordinates respectively, and results in a change in the problem parameter $\delta e_{ij}$ in Eq. (\ref{eq:ProbForm}).
A box uncertainty of $\xi=4\%$ affects every location, as shown in Fig. \ref{Im: box of 10}.
The performance comparisons shown in Fig. \ref{Im: ResultsLoc4} show that both centralized methods have a lower optimality gap than the decentralized approach, and the nominal tree provides an advantage to the CFM.
\begin{figure}[t]
    \centering
    \vspace{1.2mm}
  \subfloat[Payload capacity of one robot changed to 8 \label{Im: Q8}]{%
        \includegraphics[trim =10mm 0mm 12mm 1.5mm, clip, width=0.95\linewidth]{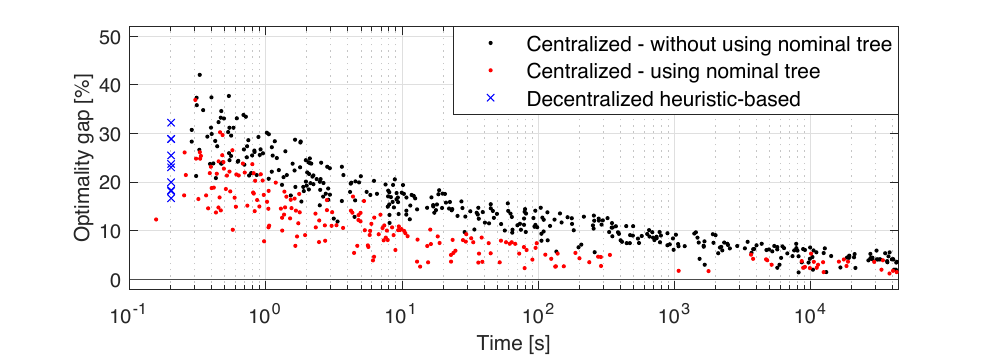}}
        \vspace{-2mm}
        \\
  \subfloat[Payload capacity of one robot changed to 6 \label{Im: Q5}]{%
        \includegraphics[trim =10mm 0mm 12mm 3.3mm, clip, width=0.95\linewidth]{Images/fPerturbedPayload_8.pdf}}
  \caption{Performance comparison for payload capacity variations}
  \vspace{-6mm}
  \label{Im:convergenceComparison:pay}
\end{figure}

\subsection{Variation in payload capacity}

Cases studied in Fig. \ref{Im:convergenceComparison:pay} relate to changes in the payload capacity of one of the robots.
It is seen that the centralized approach that uses the nominal tree is able to improve upon the decentralized solutions within 10 seconds, unlike the approach that does not use the nominal tree.

 \subsection{Discussion}

Nominal task assignment solutions that were obtained using the offline MCTS algorithm were within a 5\% optimality gap for the 25 repetitions.
However, when the decentralized algorithm used these nominal solutions to adapt to a perturbation, there was significantly higher variation in the solutions found. 
This was in addition to a higher optimality gap as compared to the centralized approaches that fully utilize the other robots of the fleet.
When the payload capacity was changed, the proposed centralized method found significantly improved solutions as compared to when the nominal tree was not used.
When a change occurs in battery capacity or pickup-delivery locations, this improvement is not as significant.
This performance variation is as expected because small changes to $\mathcal{NP}$-hard problems can result in drastic changes in the optimal solution.
In all cases and at any instance, solutions obtained from the proposed method had a lower optimality gap than when the nominal tree was not used.

\section{CONCLUSIONS}
This paper presented a centralized fleet management strategy that utilizes prior knowledge of the search space when there is a change in the nominal task definition.
The nominal material handling problem is first solved offline using an MCTS algorithm for the task assignment problem, and using a heuristic for the routing sub-problem.
When the problem is perturbed, the proposed online method evaluates the lowest-cost leaf nodes of the search tree first, rapidly producing feasible low-cost solutions.
The approach is verified to be real-time capable and is shown to perform better than computing without using the search tree and also the decentralized approach that does not attempt task reassignment.
Future work will seek to define the magnitude of perturbations that can be handled by the developed algorithm, and also its capability for larger fleets and other combinatorial problems such as the vehicle routing problem with time windows.

\bibliographystyle{IEEEtran}
\bibliography{sources}

\vfill

\end{document}